%% file: example_paper.tex
\documentclass{article}
\input{lib/my_instructions}

\usepackage{microtype}
\usepackage{graphicx}
\usepackage{subfigure}
\usepackage{booktabs} 

\usepackage{hyperref}

\usepackage[accepted]{icml2025}

\usepackage{amsmath}
\usepackage{amssymb}
\usepackage{mathtools}
\usepackage{amsthm}

\usepackage[capitalize,noabbrev]{cleveref}

\theoremstyle{plain}

\theoremstyle{definition}

\theoremstyle{remark}

\usepackage[textsize=tiny]{todonotes}
\hypersetup{
	colorlinks=true,
	linkcolor=red,
}

\icmltitlerunning{SparseVLM: Visual Token Sparsification for Efficient Vision-Language Model Inference}

\begin{document}

\twocolumn[
\icmltitle{SparseVLM: Visual Token Sparsification for Efficient \\ Vision-Language Model Inference}



\icmlsetsymbol{equal}{*}

\begin{icmlauthorlist}
\icmlauthor{Yuan Zhang}{equal,pku}
\icmlauthor{Chun-Kai Fan}{equal,pku}
\icmlauthor{Junpeng Ma}{equal,fdu}
\icmlauthor{Wenzhao Zheng}{ucb}
\icmlauthor{Tao Huang}{syu}
\icmlauthor{Kuan Cheng}{pku}
\icmlauthor{Denis Gudovskiy}{pansonic}
\icmlauthor{Tomoyuki Okuno}{pansonic}
\icmlauthor{Yohei Nakata}{pansonic}
\icmlauthor{Kurt Keutzer}{ucb}
\icmlauthor{Shanghang Zhang}{pku}
\end{icmlauthorlist}

\icmlaffiliation{pku}{State Key Laboratory of Multimedia Information Processing, School of Computer Science, Peking University}
\icmlaffiliation{ucb}{EECS, UC Berkeley}
\icmlaffiliation{fdu}{Fudan University}
\icmlaffiliation{syu}{Shanghai Jiao Tong University}
\icmlaffiliation{pansonic}{Panasonic Holdings Corporation}

\icmlcorrespondingauthor{Wenzhao Zheng}{wzzheng@berkeley.edu}
\icmlcorrespondingauthor{Shanghang Zhang}{shanghang@pku.edu.cn}

\icmlkeywords{Machine Learning, ICML}

\vskip 0.3in
]



\printAffiliationsAndNotice{\icmlEqualContribution} 

\input{sec/0_abstract}

\input{sec/1_introduction}

\input{sec/2_related_work}

\input{sec/3_method}

\input{sec/4_experiments}

\input{sec/5_analysis}

\input{sec/6_conclusion}

\section*{Acknowledgments}
This work was supported by the National Science and Technology Major Project (No. 2022ZD0117800) and by the National Natural Science Foundation of China under Grant 62472008.

\section*{Impact Statement}

Our SparseVLM provides practical advantages for deploying off-the-shelf large vision-language models on edge devices and cloud platforms. While our work does not present any evident societal implications, we believe it is unnecessary to emphasize this aspect in the current context.

\bibliography{main}
\bibliographystyle{icml2025}

\newpage
\appendix
\onecolumn

\input{sec/Appendix}

\end{document}

%% file: lib/my_instructions.tex
\input{lib/math_commands.tex}

\usepackage{colortbl}
\usepackage{xcolor}
\definecolor{mygray}{gray}{.9}
\definecolor{goldenrod}{RGB}{245,245,220}
\newlength\savewidth\newcommand\shline{\noalign{\global\savewidth\arrayrulewidth\global\arrayrulewidth 1pt}\hline\noalign{\global\arrayrulewidth\savewidth}}
\newcolumntype{a}{>{\columncolor{mygray}}c}
\usepackage{fontawesome5}
\usepackage{booktabs}
\usepackage{makecell}
\usepackage{fontawesome5}
\usepackage{lipsum}
\usepackage{comment}
\usepackage{multirow}
\usepackage{wrapfig}
\usepackage{algorithm}
\usepackage{algorithmic}
\usepackage{xfrac}  
\renewcommand{\algorithmicrequire}{\textbf{Input:}} 
\renewcommand{\algorithmicensure}{\textbf{Output:}} 

\usepackage{graphicx}
\usepackage{color}

\usepackage{amsthm}
\definecolor{darkgreen}{rgb}{0,0.7,0}

\definecolor{mygraytext}{gray}{.75}

\def\eg{\emph{e.g.}}

\def\wrt{{w.r.t.\ }}


%% file: lib/math_commands.tex
\usepackage{amsmath,amsfonts,bm}

\def\eqref#1{equation~\ref{#1}}

\def\1{\bm{1}}

\def\vp{{\bm{p}}}

\def\vr{{\bm{r}}}

\def\vx{{\bm{x}}}

\def\mA{{\bm{A}}}

\def\mH{{\bm{H}}}
\def\mI{{\bm{I}}}

\def\mK{{\bm{K}}}

\def\mO{{\bm{O}}}
\def\mP{{\bm{P}}}
\def\mQ{{\bm{Q}}}

\def\mS{{\bm{S}}}
\def\mT{{\bm{T}}}

\def\mV{{\bm{V}}}
\def\mW{{\bm{W}}}

\def\mZ{{\bm{Z}}}

\DeclareMathAlphabet{\mathsfit}{\encodingdefault}{\sfdefault}{m}{sl}
\SetMathAlphabet{\mathsfit}{bold}{\encodingdefault}{\sfdefault}{bx}{n}

\def\gK{{\mathcal{K}}}

\def\sI{{\mathbb{I}}}

\def\sL{{\mathbb{L}}}

\def\sT{{\mathbb{T}}}



%% file: sec/0_abstract.tex
\begin{abstract}
In vision-language models (VLMs), visual tokens usually bear a significant amount of computational overhead despite sparsity of information in them when compared to text tokens. 
To address this, most existing methods learn a network to prune redundant visual tokens using certain training data. Differently, we propose a text-guided training-free token optimization mechanism dubbed SparseVLM that eliminates the need of extra parameters or fine-tuning costs. Given that visual tokens complement text tokens in VLM's linguistic reasoning, we select relevant text tokens to rate the significance of visual tokens using self-attention matrices and, then, prune visual tokens using the proposed strategy to maximize sparsity while retaining information. In particular, we introduce a rank-based strategy to adaptively determine the sparsification ratio for each layer, alongside a token recycling method that compresses pruned tokens into more compact representations. Experimental results show that SparseVLM increases the efficiency of various VLMs in a number of image and video understanding tasks. 
Our code is available at \url{https://github.com/Gumpest/SparseVLMs}.
\end{abstract}

%% file: sec/1_introduction.tex
\section{Introduction}

\begin{figure*}[t]
    \centering
    \includegraphics[width=.95\textwidth]{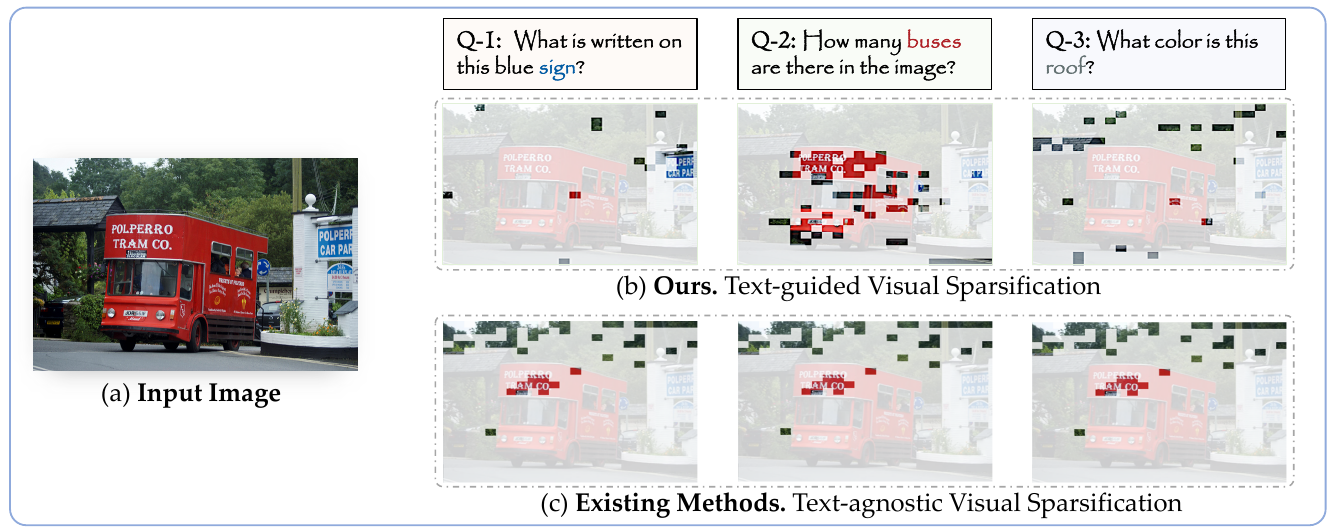}
    \caption{\textbf{Comparison of visual token sparsification methods.} Unlike previous methods with text-agnostic visual sparsification (c) e.g., VocoLLaMA \citep{ye2024voco}, our SparseVLM (b) is guided by question prompts to select relevant visual patches from the image (a).}
    \label{fig:pic1}
\end{figure*}

Benefiting from advancements in large language models (LLMs) \citep{radford2019language, brown2020language, touvron2023llama, peng2023instruction, zhang2024unveiling}, the realm of vision-language models (VLMs) has undergone significant progress. To combine visual signals with textual semantics, the mainstream practice in VLMs \citep{team2023gemini, Qwen-VL, chen2023internvl, li2024mini, li2023blip} employs sequential visual representation, where images are extracted into visual tokens and sent into an LLM decoder. With modal alignment and instruction fine-tuning \citep{du2021glm, liu2023improvedllava, zhu2023minigpt}, recent VLMs successfully adapt LLMs to the vision domain and inherit their perception and reasoning abilities.

Despite the promising performance, further incorporation of visual tokens inevitably introduces a huge memory and computational overhead when compared to LLMs, particularly for high-resolution images \citep{li2024mini} and long videos \citep{lin2023video}. For instance, a $672 \times 672$ image in LLaVA \citep{liu2024visual} yields $2304$ visual tokens that span over half of the context length. However, the information in images is typically more sparse than in natural languages \citep{marr2010vision}, resulting in inefficiency when na\"ively processing both modalities.
To address this, existing methods extract more compact image representations by modifying the image encoder or projector \citep{alayrac2022flamingo, li2023llama, instructblip, cha2024honeybee}. While some recent works further sparsify visual tokens during the decoding \citep{ye2024voco, chen2024image, shang2024llava}, they still ignore the guidance from the language tokens, which contradicts the multimodality paradigm. We argue that \textbf{\textit{visual tokens should be sparsified adaptively based on the question prompt}}, as the model might focus on different parts (e.g., foreground or background) when dealing with various questions as shown in Figure \ref{fig:pic1}. Furthermore, current approaches generally train a network to prune redundant visual tokens and require additional training data \citep{li2023llama, ye2024voco, cai2024matryoshka}.

In this paper, we introduce a text-guided training-free framework dubbed \textbf{SparseVLM} for efficient vision language model inference. 
We reuse the self-attention matrix of visual-text tokens directly from the decoder layers without extra training parameters for sparsification. 
We ascertain that \textbf{\textit{not all prompt tokens should be considered}} as some could be less relevant, which leads to inaccurate correlation results and downgrades the performance of sparse inference. 
Specifically, our SparseVLM first identifies text tokens strongly correlated with visual signals via cross-attention.
Then, we measure the contribution of visual tokens to the selected visual-relevant text tokens (i.e., ``\textit{raters}'') and adaptively prune the insignificant visual tokens. 
Instead of directly discarding the pruned tokens, we further recycle and cluster them to reconstruct more compact tokens to minimize the loss of information.
Due to the information density varying for different image inputs, we employ the rank of the attention matrix to indicate the redundancy level and set an adaptive sparsification ratio accordingly.

The proposed method is simple yet practical. It can act as a plug-and-play module to improve the efficiency of VLMs without additional fine-tuning.
Extensive experiments demonstrate that our SparseVLM effectively reduces computational overhead of various VLMs without sacrificing their performance in a wide range of image and video understanding tasks. 
For instance, LLaVA \citep{liu2024visual} when armed with SparseVLM achieves a $4.5\times$ compression rate while maintaining $97\%$ of its original performance. Alternatively, the CUDA latency can decrease by $37\%$ with only a $0.9\%$ drop in accuracy.
To investigate the effectiveness of our method in video tasks, we further apply SparseVLM to VideoLLaVA \citep{lin2023video} to compress frames with temporal dimension.
Without complex design changes, SparseVLM can sparsify video frames into an adaptive number of visual tokens and outperform existing methods in video question-answering benchmarks. Our approach consistently outperforms prior state-of-the-art FastV method \citep{chen2024image} by $11.2 - 17.3\%$ on LLaVA, $9.2 - 20.4\%$ on MiniGemini, and $14.7\%$ on VideoLLaVA when both have similar latencies.

Our main contributions are summarized as follows:
\begin{itemize}
\item We introduce a novel sparsification framework dubbed SparseVLM. To the best of our knowledge, it is the first training-free approach that explores text-aware guidance for efficient VLM inference.
\item Particularly, we propose a strategy to select relevant text tokens as raters of visual tokens, a method to assess the significance of visual tokens followed by pruning of redundant visual tokens with a recycling mechanism to minimize the loss of information.
\item When applied to a number of VLMs, SparseVLM consistently outperforms prior state-of-the-art methods in various image and video understanding benchmarks.
\end{itemize} 

%% file: sec/2_related_work.tex
\section{Related Work}

\textbf{Vision-Language Models.} Recent works on vision-language models \citep{liu2023improvedllava, chen2023internvl, li2024mini} improve multimodal comprehension and generation by processing longer visual token sequences. Moreover, the usage of higher-resolution images inevitably entails an exponential growth in the length of visual sequences. For example, LLaVA typically encodes $336 \times 336$ images into $576$ tokens \citep{liu2024visual} with up to $672 \times 672$ maximum resolution using $2880$ token sequences \citep{liu2023improvedllava}. Similarly, mini-Gemini-HD \citep{li2024mini} converts $1536 \times 1536$ high resolution and $672 \times 672$ low resolution images into $2880$ visual tokens. Moreover, comprehending videos or multiple images leads to increased token allocations for visual signals. For instance, the VideoLLaVA \citep{lin2023video} and VideoPoet \citep{kondratyuk2023videopoet} use thousands of tokens to encode multiple image frames. However, large number of visual tokens results in a computational bottleneck. Further research on sparsification is urged to further unleash VLM capabilities.

\textbf{Visual Compression for VLMs.} Compression of visual tokens is necessary because, on the one hand, their quantity is usually tens to hundreds of times that of language tokens. On the other hand, visual signals are inherently more sparse in information when compared to texts that have been produced by humans \citep{marr2010vision}. Past efforts to address the above problem can be categorized into two directions. The first one centers on the compression of a vision tower or an efficient projection of vision modality. For instance, LLaMA-VID \citep{li2023llama} exploits the Q-Former with the context token while DeCo \citep{yao2024deco} employs an adaptive pooling to downsample the visual tokens at the patch level. Methods that belong to the second direction \citep{ye2024voco, chen2024image, wu2024videollm} go deeper into the text modality and sparsify visual tokens during the LLM decoding stage, but they still lack guidance from the text tokens. In this paper, SparseVLM takes note of this limitation and improves performance upon it.


%% file: sec/3_method.tex
\section{Method}

\begin{figure*}[t]
    \centering
    \includegraphics[width=1.\textwidth]{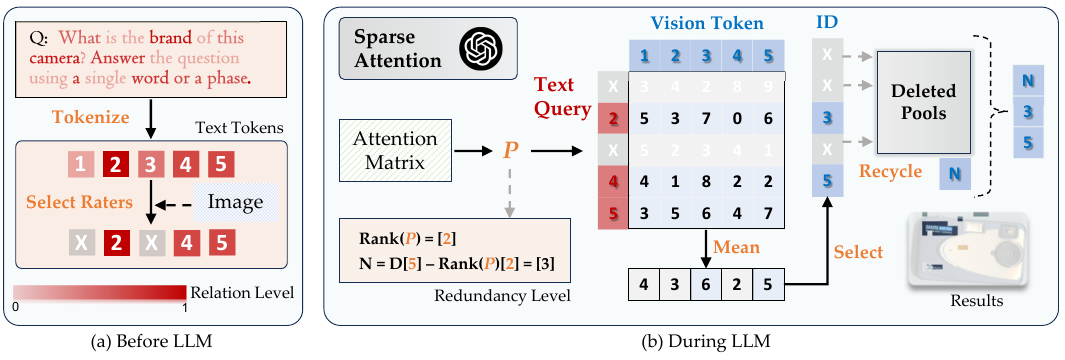}
    \caption{\textbf{The architecture of SparseVLM.} In stage (a), text raters are pre-selected before entering the sparsification LLM. In stage (b), adaptive sparsification is performed on LLM layers, involving computing redundancy and the recycling of reconstructed tokens.}
    \label{fig:pic2}
\end{figure*}

In this section, we present our SparseVLM for efficient VLM inference.
We first review the attention mechanism in VLMs and then introduce the detailed strategies for our visual sparsification including visual significance estimation, relevant text token selection, and sparsification level adaptation.
We further propose token recycling to reduce information loss and provide a theoretical analysis of computation savings.
The pipeline is shown in Figure \ref{fig:pic2}.

\subsection{Preliminary: Attention in VLM Decoders}
VLM decoders typically rely on the causal \textit{self-attention} from the original transformer architecture \citep{vaswani2017attention} for token interactions.
Without loss of generality, we describe the single-head attention below. 
Formally, the self-attention matrix with logits $\mA \in \mathbb{R}^{L\times L}$, where $L$ denotes the length of a sequence with all kinds of tokens \eg~text and visual, is computed by
\begin{equation} \label{eq:1}
    \mA = \text{Attention}(\mQ, \mK) = \textrm{Softmax} \left( \frac{\mQ \mK^T}{\sqrt{D}} \right),
\end{equation}
where the scalar $D$ represents the matrix dimension, and the $\mQ \in \mathbb{R}^{L\times D}$ and $\mK \in \mathbb{R}^{L\times D}$ are the query and key matrices, respectively. The keys and queries in a self-attention layer are computed in parallel by using multi-layer perceptrons to transform the input hidden states $\mH$ into a common space, where aligned interactions between modalities occur. 

Often, the matrix $\mA$ cannot be directly accessed due to FlashAttention-type \citep{dao2022flashattention} optimizations. Therefore, we develop an approach to extract $\mA$ while maintaining compatibility with the FlashAttention when applying our sparsification. Please refer to the Appendix \ref{a:sparseflashattn}.

\subsection{Sparsification Guidance from Text to Vision}

\textbf{Estimation of Visual Token Significance.} For a multimodal model, we aim to estimate an impact of deleting a single token from one modality to other modalities. In the VLM case, we need to quantify how relevant a visual token is to text tokens in order to determine whether it can be pruned. Therefore, we naturally reuse the self-attention logits from VLM's transformer layers as a reference since they already contain \textit{language-to-vision} query results. 

In particular, we take the interaction between the \textit{query}-dimensional part of the language modality and the \textit{key}-dimensional part of the vision modality as the basis for sparsification priority matrix $\mP \in \mathbb{R}^{L_t\times L_v}$, where $L_t$ and $L_v$ are the lengths of text and visual tokens, defined by
\begin{equation} \label{eq:2}
    \mP = \mA_{i, j},~ \textrm{and}~(i, j) \in \{ \sL, \sI \}, 
\end{equation}
where $\sL$ and $\sI$ denote the language instruction and image token sets, respectively.

Next, we obtain a vector $\tilde{\vp}$ that estimates the significance of all visual tokens \wrt the text dimension as 
\begin{equation} \label{eq:4}
    \tilde{\vp} = \left[ \tilde{p}_1, \tilde{p}_2, \ldots \tilde{p}_{L_v} \right] = \frac{1}{L_t} \sum\nolimits_{i=1}^{L_t} \mP_{i},
\end{equation}
where we use $\tilde{\vp}$ as an indicator for sparsification and a larger value in $\tilde{\vp}$ means higher significance of the corresponding visual token.
Calculation of (\ref{eq:4}) costs $L_t \times L_v$ FLOPs only while the access to already computed $\mA$ is considered as free, which is highlights low complexity of the SparseVLM.

\begin{figure*}[t]
    \centering
    \includegraphics[width=1.\textwidth]{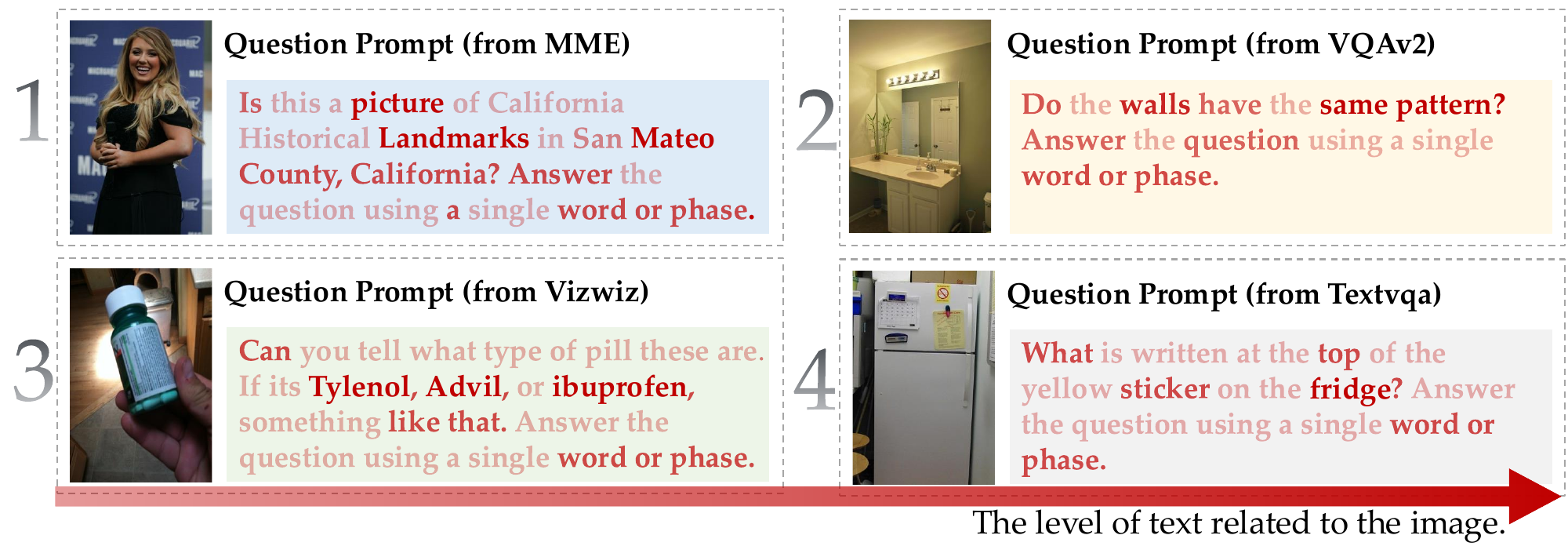}
    \vspace{-6mm}
    \caption{\textbf{Sample prompts from four representative multimodal benchmarks.} The darker the word, the greater its relationship to the image and the more valuable it is for reference. 
    We see that some words are irrelevant to the vision domain (e.g., prepositions and pronouns) and should not be considered for visual sparsification. It is best viewed in color.}
    \vspace{-2mm}
    \label{fig:pic3}
\end{figure*}

\textbf{Relevant Text Token Selection.} It is not appropriate to use all text tokens as a reference for visual sparsification. Figure \ref{fig:pic3} shows four representative cases where we compute the correlation between the prompt and the image. Case $3$ highlights \verb+Tylenol, Advil, ibuprofen+, while \verb+sticker, fridge+ in case $4$ are significant, where a large proportion of question tokens in light red include little visual relevance. Therefore, it is unreasonable to make insignificant text tokens to rate visual tokens, and we need to select relevant text tokens (i.e., ``\textit{raters}'') for guidance.

Specifically, for an input image $\vx_v$, the vision embedding tokens $\mH_v$ can be computed as
\begin{equation} \label{eq:5}
    \mH_v = \mW \mZ_v,
\end{equation}
where $\mZ_v$ is the visual feature provided by visual encoder $\mZ_v = g(\vx_v)$, and $\mW$ is the projection matrix to convert $\mZ_v$ into vision embedding tokens $\mH_v$. 
For the language instruction $\vx_q$, it is transformed into text embedding tokens $\mH_q$ through the tokenizer. The above tokens both have the same dimensionality as the word embedding space. Then, we start to recognize which characters in the prompt are visually relevant and assign them the role of raters, which can be formulated as
\begin{equation} \label{eq:6}
    \bm{s} = \{\ i \ | \ \vr_i \geq m \},  \ \ i \in \{1, 2, \ldots, L_t \},
\end{equation}
\begin{equation} \label{eq:7}
    \vr = \frac{1}{L_v} \sum_{j=1}^{L_v} \left( \textrm{Softmax} \left( \mH_v {\mH_q}^T \right) \right)_j,
\end{equation}
where $m = \textrm{mean}(\vr)$ and only candidates that exceed the $m$ threshold become raters. The strategy $\bm{s}$ contains the indices of selected raters from the candidate list of $L_t$ tokens. The (\ref{eq:7}) costs $L_t \times L_v \times 2D$ FLOPs that is only computed once before the decoder layer processing.

\textbf{Sparsification Level Adaptation.} 
Having obtained the token significance, we further propose a rank-based strategy to adaptively determine the level of vision sparsification at each decoder layer.
Considering that \textbf{\textit{a full-rank matrix implies that all its rows or columns are linearly independent}}, we use the rank of $\mP$ to demonstrate the redundancy of the visual tokens. 
We argue that the difference between the dimension and rank of $\mP$ reflects its redundancy and utilize a scaling factor $\lambda$ to determine the number of deletions as
\begin{equation} \label{eq:8}
    N = \lambda \times (L_v - \textrm{rank}(\mP)).
\end{equation}
We then remove $N$ visual tokens with the smallest values in $\mP$. 
Notably, if the result of $N$ in a decoder layer is $0$, we skip the layer without sparsification. 
This stage requires $L_t \times L_v \times \text{min}(L_t, L_v)$ FLOPs for rank computation.

\subsection{Visual Token Recycling} 
We progressively sparsify visual tokens in each layer in the decoder, which results in more discarded tokens at later stages. 
Despite being less significant, the pruned visual tokens with relatively large values in $\mP$ still contain certain information. 
To efficiently preserve more visual details with fewer tokens, we propose a token recycling strategy to aggregate and reconstruct tokens to be pruned. 

\textbf{Token Aggregation.} 
We first recycle the pruned visual tokens $\bar{h}_v$ with the top-$\tau$ (\%) highest values in $\mP$ from the deleted pool. 
Then, we group $\bar{h}_v$ tokens with $k$-nearest neighbor density peak aggregation algorithm \citep{rodriguez2014clustering} for adaptive token aggregation. 

In particular, we first compute the local density $\rho_i$ of the $i$th token of total $\tau \times N$ recycled tokens according to its $k$-nearest neighbors $\gK (\bar{h}_{v}^{i})$ as
\begin{equation}
\rho_{i}=\exp \left(-\frac{1}{k} \sum\nolimits_{\bar{h}_{v}^{j} \in \gK (\bar{h}_{v}^{i})}^{i, j} \left\|\bar{h}_{v}^{i}-\bar{h}_{v}^{j}\right\|_{2}^{2} \right).
\label{eq:9}
\end{equation}
Then, we compute the minimum distance between the recycled token $\bar{h}_{v}^{i}$ and any other token with higher density (denoted as the distance indicator $\delta_{i}$) that is defined by
\begin{equation}
\delta_{i} = \left\{
\begin{array}{ll}
  \min & \left\|\bar{h}_{v}^{i}-\bar{h}_{v}^{j} \right\|_{2}, \text { if } \exists j \text { s.t. } \rho_{j}>\rho_{i}, \\
  \max & \left\|\bar{h}_{v}^{i}-\bar{h}_{v}^{j}\right\|_{2}, \text { otherwise }.
\end{array}
\right.
\label{eq:10}
\end{equation}
We use $\rho_i \times \delta_i$ to indicate the score of each token, where the tokens with higher scores are likely to be cluster centers.
Other tokens are then assigned to the nearest cluster center via cosine similarity. 
The FLOPs cost in this stage is $L_r\times(3L_r - 1) \times 2D + L_{r}$, where $L_{r} = \tau \times N$ is the length of recycled tokens, $C = \theta \times L_{r} $ is the number of cluster centers, and $\tau$ and $\theta$ are hyperparameters.

\textbf{Token Reconstruction.} 
Having performed token aggregation, the recycled tokens with similar semantics are classified into the same group. 
Then, the tokens $\sT \in \mathbb{R}^{N_k \times D}$ in the $k$th group are reconstructed into a new compressed token $\mT_{k} \in \mathbb{R}^{1 \times D}$ via the element-wise sum operation as
\begin{equation}
\mT_{k} = \sum\nolimits_{i=1}^{N_k} \sT[i], \ \ k \in \{1, 2, \ldots, C\},
\label{eq:11}
\end{equation}
where $N_k$ is the token number of the $k$th group and the operation costs $D \times (L_{r} - C)$ FLOPs.

\subsection{Theoretical Analysis of Computational Complexity}
We consider the computation of multi-head attention and feed-forward network (FFN) modules in the FLOPs estimation. 
Assuming $N$ is the number of pruned tokens, $D$ is the hidden state size, which is the same as the intermediate size in FFN, the FLOPs for one Transformer layer can be reduced by $6 (N - C) D^{2} + 2(N - C)^{2} D$. 
Besides, our partial step introduces minimal computation with the details provided in Appendix \ref{a:flops}.
Thus, we estimate the FLOPs savings as the reduction part minus the additional overhead:
\begin{equation}
\begin{array}{ll}
 \underbrace{\scriptstyle { \textstyle \sum_{i}} 6 (N_{i} - C_{i}) D^{2} + 2 (N_{i} - C_{i})^{2} D}_{\text{reduction part}}
 - \\
 \underbrace{\scriptstyle 2 L_{t} L_{v} D - {\textstyle \sum_{i}} L_{t}^{i} L_{v}^{i} (1 + \min (L_{t}^{i}, L_{v}^{i})) -(6{L^{i}_{r}}^{2} + 2L_{r}^{i}) D - L_{r}^{i}}_{\text{overhead part}} \\
 \approx \scriptstyle -2 L_{t} L_{v} D + \sum_{i} D(6DN_{i}(1-x) + N_{i}^{2}(2+2x^2-4x-6(\tau)^{2})) - {L^{i}_{t}}^{2} L_{v}^{i} \\
 \approx \scriptstyle -2 L_{t} L_{v} D + \sum_{i} DN_{i}(6D + 2N_{i}) - {L^{i}_{t}}^{2} L_{v}^{i},
\label{eq:12}
\end{array}
\end{equation}
where $i \in \{1, 2, \ldots, \Omega\}$ and $\Omega$ is the number of total layers, and $x = \tau \times \theta$ is a very small decimal that can be ignored. 

%% file: sec/4_experiments.tex
\section{Experiments}

\input{table/main}

In this section, we validate our method within various vision-language architectures on comprehensive multimodal benchmarks, including image and video understanding tasks, to assess its generality, effectiveness, and efficiency.

\subsection{Image Understanding Tasks}
\textbf{Datasets.} For image-based multimodal evaluation, we conduct experiments on eight widely adopted benchmarks, including GQA \citep{hudson2019gqa}, MMBench (MMB) \citep{liu2023mmbench}, MME \citep{fu2023mme}, POPE \citep{li2023evaluating}, SQA \citep{lu2022learn}, SEED-Bench (SEED) \citep{li2024seed}, VQA$^{\text{Text}}$ (TextVQA) \citep{singh2019towards}, and MMVet \citep{yu2023mm}.

\textbf{Implementation Details.} We verify SparseVLM on three VLM frameworks: LLaVA \citep{liu2024visual}, Mini-Gemini (MGM) \citep{li2024mini}, and Qwen2-VL \citep{Qwen-VL}. LLaVA-1.5 employs CLIP-pretrained ViT-L as the visual tower, MGM further introduces a LAION-pretrained ConvNeXt-L \citep{liu2022convnet} for high-resolution refinement, while Qwen2-VL owns dynamic resolution encoder.

\textbf{Main Results.} In Table \ref{tab:main}, we present the performance of SparseLLaVA (LLaVA equipped with SparseVLM) on image understanding benchmarks. To intuitively assess the performance, we provide the results by percentage format for comparative analysis, and the accuracy of the vanilla model with the 100\% upper limit. We set $3$ vision token count configurations ($192$, $128$, and $64$) to check the advantages of SparseVLM comprehensively. When pruning from $576$ to $192$ tokens, the SparseLLaVA only decreases the average accuracy by $0.9\%$ without additional training and exceeds ToMe \citep{bolya2022token} $10.2\%$. When only $64$ tokens are kept, our method outperforms FastV \citep{chen2024image} by a significant margin of \textbf{17.3}\%, while ToMe performs worst due to its direct merging. Furthermore, we also compare the recent method PDrop \citep{xing2024pyramiddrop} training-free version, which has lower FLOPs computation. However, our method outperforms it in accuracy and latency, which are the most crucial metrics for practical deployment.

\begin{figure*}[t]
    \centering
    \includegraphics[width=.97\textwidth]{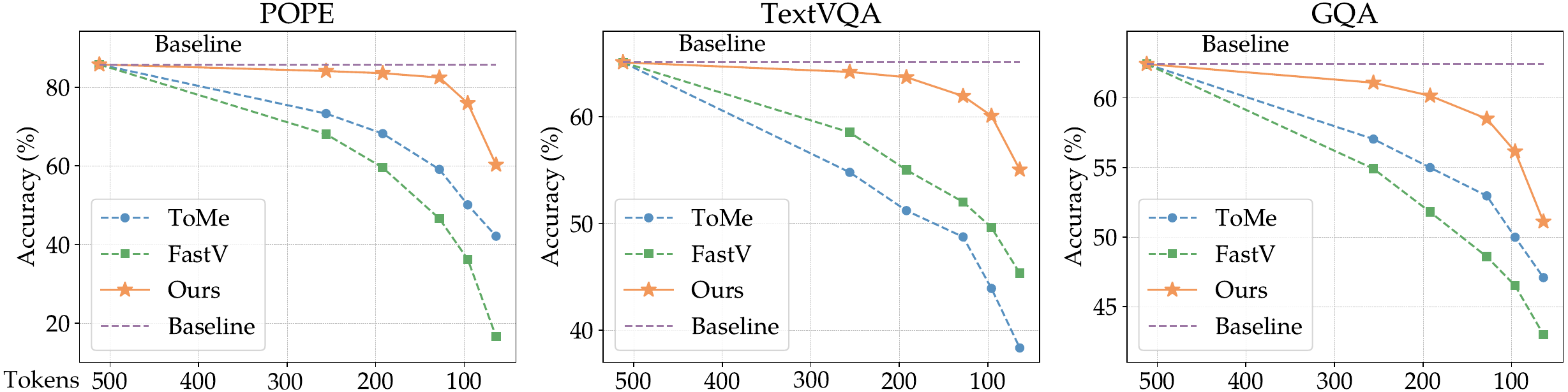}
    \vspace{-2mm}
    \caption{\textbf{Performance of MGM w/ SparseVLM on three multimodal benchmarks.} The horizontal axis represents the remaining number of vision tokens, while the vertical axis means the accuracy after percentage normalization.}
    \label{fig:mgm}
    \vspace{-3mm}
\end{figure*}

Figure \ref{fig:mgm} visualizes the performance of SparseMGM on POPE, TextVQA, and GQA. We find that our framework has an obvious advantage over FastV and ToMe. With the reduction of tokens, the gap between FastV and SparseVLM is increasing sharply. The reason is that, compared to FastV and ToMe, the text-aware strategy enables us to accurately locate visual tokens with more details, while the recycling of pruned tokens further reduces information loss.

\input{table/qwen2-vl}
We further investigate our efficacy on Qwen2-VL. In Table \ref{tab:qwen2-vl}, when $54.5\%$ of vision tokens are removed, Qwen2-VL maintains an accuracy of $98.0\%$. Furthermore, for every 100 tokens pruned, the accuracy only drops by approximately $0.8\%$. This validates the effectiveness of our method at high resolutions and its compatibility with variable resolutions.

\subsection{Video Understanding Tasks}
\input{table/videollava}
\textbf{Datasets.} We test on four common video question answering benchmarks, TGIF-QA \citep{jang2017tgif}, MSVD-QA \citep{xu2017video}, MSRVTT-QA \citep{xu2017video}, and ActivityNet-QA \citep{yu2019activityqa}. Specifically, following FastV's \citep{chen2024image} setup, we use the first 1000 samples per benchmark and score them using the Video-ChatGPT \citep{maaz2023video} evaluation tool, acknowledging the characteristic length imbalances in these datasets.

\textbf{Implementation Details.}
We directly apply our SparseVLM for Video-LLaVA \citep{lin2023video}, which is composed of several key components, including language bind encoder $f^{v}_{M} $\citep{zhu2023languagebind} for extracting features from raw visual inputs (e.g., images or videos), a language decoder model $f_L$ such as Vicuna \citep{touvron2023llama}, a visual projection layer $f_P$, and a word embedding layer $f_T$.

\textbf{Main Results.}
In Table \ref{tab:videollava}, we set the Video-LLaVA with $2048$ video tokens as our upper bound for an overall average accuracy of $100.0\%$ and a score of $+0.00$. To make a fair comparison, we both preserve $194$ vision tokens ($90.5\%$ pruning ratio) for FastV \citep{chen2024image} and SparseVLM. 
It is clear that our approach consistently outperforms FastV across all benchmarks, both in accuracy (Acc.) and GPT evaluation score. SparseVideoLLaVA achieves a total average accuracy of $95.0\%$, a significant $\textbf{14.7\%}$ higher than $80.3\%$ of FastV. (From the GPT score perspective, SparseVLM only loses $0.04$ points compared to $0.17$ points of FastV.)
These improvements suggest that when handling video modality containing temporal features, SparseVLM continues to deliver strong performance, generating accurate responses to diverse questions while utilizing significantly fewer tokens. This achieves an effective trade-off between inference efficiency and model performance.

%% file: table/main.tex
\renewcommand{\multirowsetup}{\centering}
\definecolor{mygray}{gray}{.92}
\definecolor{ForestGreen}{RGB}{34,139,34}
\newcommand{\fg}[1]{\mathbf{\mathcolor{ForestGreen}{#1}}}
\definecolor{Forestred}{RGB}{220,50,50}
\newcommand{\fr}[1]{\mathbf{\mathcolor{Forestred}{#1}}}
\begin{table*}[t]
    \centering
    \vspace{-2mm}
    \setlength{\tabcolsep}{2.8pt}
    \renewcommand{\arraystretch}{1.33}
    \footnotesize
	\centering
	\caption{\textbf{Performance of SparseLLaVA under different vision token configurations.} The vanilla number of vision tokens is $576$. The first line of each method is the raw accuracy of benchmarks, and the second line is the proportion relative to the upper limit.}
    \vspace{2mm}
	\label{tab:main}
    \begin{tabular}{p{2.1cm}|c c c c c c c c | p{1.85cm} | p{1.83cm} | p{1.85cm}}
        \shline
        \textbf{Method} & \textbf{GQA} & \textbf{MMB} & \textbf{MME} & \textbf{POPE} & \textbf{SQA} & \textbf{SEED} & \textbf{VQA}$^{\text{Text}}$ & \textbf{MMVet} &\makecell[c]{\textbf{Acc. (\%)}} & \makecell[c]{\textbf{FLOPs (T)}} & \makecell[c]{\textbf{Latency (ms)}}\\
        \shline
        \rowcolor{mygray}
        \multicolumn{12}{c}{\textit{Upper Bound, 576 Tokens} \ $\textbf{(100\%)}$}\\
        \multirow{2}*{Vanilla} & 61.9 & 64.6 & 1864 & 85.9 & 69.5 & 60.3 & 58.3 & 30.9 & \makecell[c]{\multirow{2}*{100}} & \makecell[c]{\multirow{2}*{4.62}} & \makecell[c]{\multirow{2}*{57.82}}\\ 
        ~ & 100\% & 100\% & 100\% & 100\% & 100\% & 100\% & 100\% & 100\% & ~ & ~ & ~\\
        \hline

        \rowcolor{mygray}
        \multicolumn{12}{c}{\textit{Retain 192 Tokens} \ $\fg{(\downarrow 66.7\%)}$} \\
        \multirow{2}*{ToMe \texttt{\scriptsize{(ICLR23)}}} & 54.3 & 60.5 & 1563 & 72.4 & 65.2 & 53.1 & 52.1 & 27.9 & \makecell[c]{\multirow{2}*{88.9 $(\downarrow 11.1)$}} & \makecell[c]{\multirow{2}*{2.05}} & \makecell[c]{\multirow{2}*{34.06}}\\
        ~ & 87.7\% & 93.5\% & 83.9\% & 84.3\% & 93.8\% & 88.1\% & 89.5\% & 90.3\% & ~ & ~\\
        \hline
        \multirow{2}*{FastV \texttt{\scriptsize{(ECCV24)}}} & 52.6 & 61.0 & 1605 & 64.8 & 69.1 & 52.1 & 52.5 & 26.7 & \makecell[c]{\multirow{2}*{87.9 $(\downarrow 12.1)$}} & \makecell[c]{\multirow{2}*{2.11}} & \makecell[c]{\multirow{2}*{34.87}}\\
        ~ & 85.0\% & 94.4\% & 86.1\% & 75.4\% & 99.4\% & 86.4\% & 90.1\% & 86.4\% & ~ & ~\\
        \hline
        \multirow{2}*{PDrop \texttt{\scriptsize{(CVPR25)}}} & 57.1 & 63.2 & 1766 & 82.3 & 70.2 & 54.7 & 56.1 & 30.5 & \makecell[c]{\multirow{2}*{95.9 $(\downarrow 4.1)$}} & \makecell[c]{\multirow{2}*{2.03}} & \makecell[c]{\multirow{2}*{36.74}}\\
        ~ & 92.2\% & 97.8\% & 94.7\% & 95.8\% & 101.0\% & 90.7\% & 96.2\% & 98.7\% & ~ & ~\\
        \hline
        \multirow{2}*{SparseVLM} & 59.5 & 64.1 & 1787 & 85.3 & 68.7 & 58.7 & 57.8 & 33.1 & \makecell[c]{\multirow{2}*{99.1 $\fg{(\downarrow 0.9)}$}} & \makecell[c]{\multirow{2}*{2.14}} & \makecell[c]{\multirow{2}*{36.50}}\\
        ~ & 96.1\% & 99.2\% & 95.9\% & 99.3\% & 98.8\% & 97.3\% & 99.1\% & 107.1\% & ~ & \\
        \hline

        \rowcolor{mygray}
        \multicolumn{12}{c}{\textit{Retain 128 Tokens} \ $\fg{(\downarrow 77.8\%)}$}\\
        \multirow{2}*{ToMe \texttt{\scriptsize{(ICLR23)}}} & 52.4 & 53.3 & 1343 & 62.8 & 59.6 & 50.9 & 49.1 & 27.2 & \makecell[c]{\multirow{2}*{81.9 $(\downarrow 18.1)$}} & \makecell[c]{\multirow{2}*{1.62}} & \makecell[c]{\multirow{2}*{30.00}}\\
        ~ & 84.7\% & 82.4\% & 72.1\% & 73.1\% & 85.8\% & 84.4\% & 84.4\% & 88.0\% & ~ & ~ \\
        \hline
        \multirow{2}*{FastV \texttt{\scriptsize{(ECCV24)}}} & 49.6 & 56.1 & 1490 & 53.4 & 68.6 & 48.1 & 50.5 & 26.3 & \makecell[c]{\multirow{2}*{82.4 $(\downarrow 17.6)$}} & \makecell[c]{\multirow{2}*{1.70}} & \makecell[c]{\multirow{2}*{30.70}}\\
        ~ & 80.1\% & 86.8\% & 79.9\% & 62.2\% & 98.7\% & 79.8\% & 86.6\% & 85.1\% & ~ & \\
        \hline
        \multirow{2}*{PDrop \texttt{\scriptsize{(CVPR25)}}} & 56.0 & 61.1 & 1664 & 82.3 & 69.9 & 53.3 & 55.1 & 30.8 & \makecell[c]{\multirow{2}*{94.3 $(\downarrow 5.7)$}} & \makecell[c]{\multirow{2}*{1.62}} & \makecell[c]{\multirow{2}*{37.77}}\\
        ~ & 90.5\% & 95.4\% & 89.3\% & 95.8\% & 100.6\% & 88.4\% & 94.5\% & 99.7\% & ~ & ~\\
        \hline
        \multirow{2}*{SparseVLM} & 58.4 & 64.5 & 1746 & 85.0 & 68.6 & 58.2 & 56.7 & 29.0 & \makecell[c]{\multirow{2}*{96.7 $\fg{(\downarrow 3.3)}$}} & \makecell[c]{\multirow{2}*{1.72}} & \makecell[c]{\multirow{2}*{33.28}}\\
        ~ & 94.3\% & 99.8\% & 93.7\% & 99.0\% & 98.7\% & 96.5\% & 97.3\% & 93.9\% & ~ & ~\\
        \hline

        \rowcolor{mygray}
        \multicolumn{12}{c}{\textit{Retain 64 Tokens} \ $\fg{(\downarrow 88.9\%)}$}\\
        \multirow{2}*{ToMe \texttt{\scriptsize{(ICLR23)}}} & 48.6 & 43.7 & 1138 & 52.5 & 50.0 & 44.0 & 45.3 & 24.1 & \makecell[c]{\multirow{2}*{71.1 $(\downarrow 28.9)$}} & \makecell[c]{\multirow{2}*{1.19}} & \makecell[c]{\multirow{2}*{26.52}}\\
        ~ & 78.5\% & 67.5\% & 61.1\% & 61.1\% & 71.9\% & 73.0\% & 77.8\% & 78.0\% & ~ & \\
        \hline
        \multirow{2}*{FastV \texttt{\scriptsize{(ECCV24)}}} & 46.1 & 47.2 & 1255 & 38.2 & 68.7 & 43.7 & 47.8 & 19.6 & \makecell[c]{\multirow{2}*{72.0 $(\downarrow 28.0)$}} & \makecell[c]{\multirow{2}*{1.29}} & \makecell[c]{\multirow{2}*{27.30}}\\
        ~ & 74.5\% & 73.1\% & 67.3\% & 44.5\% & 98.8\% & 72.5\% & 82.0\% & 63.4\% & ~ & \\
        \hline
        \multirow{2}*{PDrop \texttt{\scriptsize{(CVPR25)}}} & 41.9 & 33.3 & 1092 & 55.9 & 69.2 & 40.0 & 45.9 & 30.7 & \makecell[c]{\multirow{2}*{73.4 $(\downarrow 26.6)$}} & \makecell[c]{\multirow{2}*{1.18}} & \makecell[c]{\multirow{2}*{43.41}}\\
        ~ & 67.7\% & 51.6\% & 58.6\% & 65.1\% & 99.6\% & 66.3\% & 78.7\% & 99.4\% & ~ & ~\\
        \hline
        \multirow{2}*{SparseVLM} & 53.8 & 60.1 & 1589 & 77.5 & 69.8 & 52.2 & 53.4 & 24.9 & \makecell[c]{\multirow{2}*{89.3 $\fg{(\downarrow 10.7)}$}} & \makecell[c]{\multirow{2}*{1.30}} & \makecell[c]{\multirow{2}*{29.89}} \\
        ~ & 86.9\% & 93.0\% & 85.2\% & 90.2\% & 100.4\% & 86.6\% & 91.6\% & 80.6\% & ~ &\\

        \shline
	\end{tabular}
     \vspace{-2mm}
\end{table*}

%% file: table/qwen2-vl.tex
\renewcommand{\arraystretch}{1.02}
\begin{table}[t]
  \vspace{-3mm}
  \caption{\textbf{Performance of SparseVLM on Qwen2-VL.}}
    \vspace{1mm}
  \renewcommand{\arraystretch}{1.6}
  \label{tab:qwen2-vl}
  \centering
  \small
  \begin{tabular}{l|c|c|c|c}
    \toprule
    \textbf{Tokens} & \textbf{MMB} & \textbf{POPE} & \textbf{VQA}$^{\text{Text}}$ & \textbf{Avg.}\\
    \midrule
    Dynamic & 80.5 (1323) & 86.4 (1311) & 84.3 (1326) & 83.7 \\
    \midrule
    600 & 79.6 & 86.5 & 80.3 & 82.1 \\
    500 & 78.8 & 86.3 & 79.0 & 81.4 \\
    400 & 79.0 & 85.8 & 77.1 & 80.7 \\
    \bottomrule
  \end{tabular}
\vspace{-10pt} 
\end{table}

%% file: table/videollava.tex
\renewcommand{\multirowsetup}{\centering}
\definecolor{mygray}{gray}{.90}

\begin{table}[t]
    \centering
    \setlength{\tabcolsep}{2pt} 
    \renewcommand{\arraystretch}{1.85} 
    \footnotesize
    \setlength\tabcolsep{4pt}
    \vspace{-3mm}
    \caption{\textbf{The results of Video-LLaVA with SparseVLM on video question answering task.} The original number of video tokens is $2048$, while our experiment collectively prunes it down to $194$ tokens. FastV \cite{chen2024image} is included for comparison.}
    \label{tab:videollava}
    \vspace{1mm}
    \resizebox{\columnwidth}{!}{%
    \begin{tabular}{@{}l|cc|cc|cc|cc|cc@{}}
        \toprule
        \multirow{2}{*}{Method} 
          & \multicolumn{2}{c|}{\textbf{TGIF}} 
          & \multicolumn{2}{c|}{\textbf{MSVD}} 
          & \multicolumn{2}{c|}{\textbf{MSRVTT}} 
          & \multicolumn{2}{c|}{\textbf{ActivityNet}} 
          & \multicolumn{2}{c}{\textbf{Avg.}} \\[-0.2em]
        \cmidrule(lr){2-3} \cmidrule(lr){4-5} \cmidrule(lr){6-7} \cmidrule(lr){8-9} \cmidrule(lr){10-11}
          & Acc. & Score
          & Acc. & Score
          & Acc. & Score
          & Acc. & Score 
          & Acc. & Score \\
        \midrule
        \multirow{2}{*}{Video-LLaVA} & 18.9 & 2.54 & 72.0 & 3.95 & 57.1 & 3.45 & 43.6 & 3.81 & \multirow{2}*{47.9} & \multirow{2}*{3.44} \\   
        ~ & 100\% & +0.00 & 100\% & +0.00 & 100\% & +0.00 & 100\% & +0.00 ~ \\
        \midrule
        \multirow{2}{*}{FastV \texttt{\scriptsize{(ECCV24)}}} & 10.2 & 2.29 & 58.3 & 3.62 & 52.3 & 3.42 & 41.3 & 3.76 & \multirow{2}*{40.5} & \multirow{2}*{3.27} \\ 
        ~ & 54.0\% & -0.34 & 81.0\% & -0.33 & 91.6\% & -0.03 & 94.7\% & -0.12 & ~ \\
        \midrule
        \multirow{2}{*}{Ours} & 14.9 & 2.41 & 71.7 & 3.94 & 56.1 & 3.43 & 45.1 & 3.81 & \multirow{2}*{\textbf{47.0}} & \multirow{2}*{\textbf{3.40}} \\ 
        ~ & 78.8\% & -0.13 & 99.6\% & -0.01 & 98.3\% & -0.02 & 103.4\% & -0.00 & ~ \\
        \bottomrule
    \end{tabular}}
    \vspace{-4mm}
\end{table}

%% file: sec/5_analysis.tex
\section{Analysis}
\subsection{Relevant Text Token Selection}

\begin{figure}[t]
    \centering
    \includegraphics[width=0.48\textwidth]{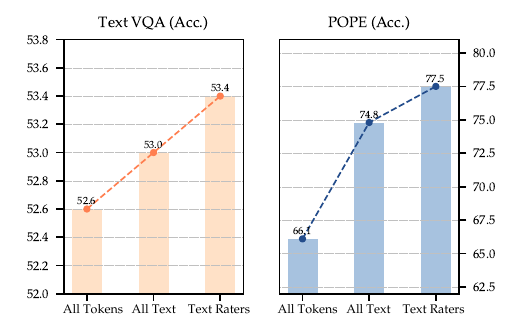}
    \vspace{-6mm}
    \caption{\textbf{The ablation study of text raters on LLaVA 7B.} }
    \vspace{-6mm}
    \label{fig:ab_1}
\end{figure}

\begin{figure*}[t]
    \centering
    \vspace{-2mm}
    \includegraphics[width=.98\textwidth]{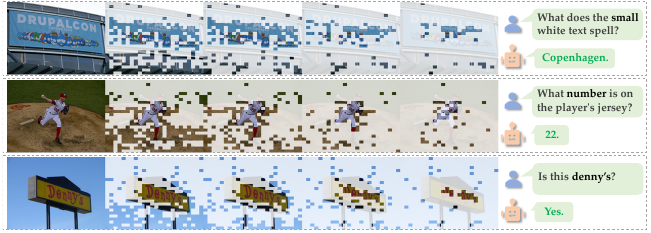}
    \vspace{-4mm}
    \caption{\textbf{Visualization of SparseVLM on different VQA prompts.} From left to right, the visual representation becomes increasingly sparse, leaving fewer vision tokens. Best viewed in color.}
    \label{fig:visualization}
    \vspace{-7mm}
\end{figure*}

We propose a selection mechanism to localize visually irrelevant text tokens to limit their negative effects in rating the significance of vision tokens. Here we conduct experiments to analyze the effects of the mechanism in Figure \ref{fig:ab_1}. Under the same number of vision tokens ($64$), we have 3 settings (using all tokens, only text tokens, and only text raters we select) with LLaVA \citep{liu2023improvedllava} to judge vision token candidates. In TextVQA \citep{singh2019towards}, by building upon the text-aware manner, our mechanism improves the baseline (all tokens) by $0.8\%$, which validates that our extra selection is effective. Besides, we further outperform the vanilla text-aware method (only text tokens) by $\textbf{2.7}\%$ on POPE \citep{li2023evaluating}. The huge margin means POPE sparsification is quite sensitive to question prompts, and text guidance is necessary. In summary, text rater selection is general and improves the performance across scenarios.

\subsection{Recycling of Pruned Tokens}
\input{table/merge_ablation}
To validate the effectiveness of our token recycling strategy, we perform ablation experiments on the LLaVA model \citep{liu2023improvedllava}. The results are presented in Table \ref{tab:merge_ablation}. Across multiple sparsity ratios ($64$, $96$, $128$, $192$), our algorithm achieves a significant average performance improvement of $\textbf{1.2}\%$ and $\textbf{7.2}\%$ on TextVQA \citep{singh2019towards} and POPE \citep{li2023evaluating}, respectively. Notably, as the number of pruned vision tokens increases, the benefit brought by our recycling method increases. For instance, when pruning from $192$ to $64$ tokens, the pruned token recycling significantly boosts the accuracy from $\textbf{1.5\%}$ to $\textbf{17.7\%}$ on POPE. We argue that when the size of the deleted pool grows, the amount of lost information increases. Our method effectively recycles the lost information and compresses it into few slots using the proposed reconstruction mechanism. 

\subsection{Computational Efficiency}
SparseVLM affords significant efficiency and storage gains for the inference process. We conduct a comparative analysis of CUDA time, and FLOPs on LLaVA-7B, and compare our method with the baseline method and FastV \citep{chen2024image}. 
As displayed in Table \ref{tab:main}, we conduct an inference efficiency analysis on a single NVIDIA A100-80GB with identical lengths of text prompts and single-image inputs. 
Compared to the baseline model, SparseVLM achieves a significant reduction of $43.1\%$ in CUDA time and $62.8\%$ in FLOPs while keeping $96.7\%$ accuracy. Despite SparseVLM has a minimal overhead to calculate text raters and cluster-pruned vision tokens, it leads to fewer than FastV tokens with comparable accuracy. Additionally, SparseVLM saves $67\%$ cache memory compared to vanilla LLaVA (where 302.4MB is reduced to 100.8MB), while keeping $99.1\%$ accuracy. More efficiency visualization (e.g., efficiency on VideoLLaVA) can be found in the Appendix \ref{a:trade-off}.

\subsection{Qualitative Visualization}
As shown in Figure \ref{fig:visualization}, we visualize SparseVLM on various VQA questions. From left to right, we visualize the results after we apply token pruning to different layers. As the number of layers increases, more tokens are pruned and the Region of Interest (ROI) is gradually refined. The model systematically reduces less relevant image information while retaining key tokens closely tied to the question. The visualization reveals that SparseVLM, although discarding some overall image details, effectively retains essential visual tokens. These preserved tokens encapsulate the features necessary for answering the question, focusing on more relevant visual regions through their interaction with the question. More cases are in the Appendix \ref{a:visualization}.

%% file: table/merge_ablation.tex
\begin{table}[t] 
    \centering
    \setlength{\tabcolsep}{8pt}
    \renewcommand{\arraystretch}{1.55}
    \footnotesize
	\caption{\textbf{Ablation study on token reconstruction (TR).} Experiments are conducted on GQA and POPE on LLaVA 7B.}
 \vspace{2mm}
	\label{tab:merge_ablation}
     \begin{tabular}
     {c | c c  c c | c }
    \toprule
     \centering
     \multirow{2}{*}{\textbf{Benchmark}} &\multicolumn{4}{c|}{\textbf{Tokens}} & \multirow{2}{*}{\textbf{Avg.}}\\
    \cline{2-5}
    ~ & \textbf{64} & \textbf{96} & \textbf{128} & \textbf{192} & \\
    \midrule
    \textbf{GQA} & 52.2 & 55.2 & 58.1 & 59.4 & 56.2 \\
    \rowcolor{mygray}
    \textbf{+} TR & \textbf{53.8}&
    \textbf{56.4}&
    \textbf{58.4}&
    \textbf{59.5}&
    \textbf{57.0}  \\
    \midrule
    
    \textbf{POPE} & 72.8 &77.5&83.7&85.2& 79.8  \\
    \rowcolor{mygray}
    \textbf{+} TR & \textbf{77.5}&
    \textbf{81.9}&
    \textbf{85.0}&
    \textbf{85.3}&
    \textbf{82.4}   \\
    \bottomrule

     \end{tabular}
     \vspace{-6mm}
\end{table}

%% file: sec/6_conclusion.tex
\section{Conclusion}
This paper introduced a text-aware training-free token optimization approach called SparseVLM which significantly decreased the test-time computations of various VLMs. Unlike prior methods, SparseVLM optimized VLMs without introducing extra parameters and fine-tuning costs.
We achieved a more compact visual representation by employing the rank of attention matrices to determine pruning ratios and by recycling the pruned tokens via the reconstruction mechanism to reduce the information loss.
Experiments demonstrated that \eg~the LLaVA when equipped with SparseVLM achieved 37.0\% reduction in latency with a compression ratio of 77.8\% while maintaining 97\% of the original accuracy. 
Moreover, our method exceeded FastV accuracy by 14.7\% in video understanding tasks. Our SparseVLM can provide practical benefits for deploying off-the-shelf VLMs on edge devices and in the cloud setting.

%% file: sec/Appendix.tex
\section*{Appendix}

\section{The Redundancy of Visual Tokens in VLMs}
\begin{figure*}[htbp]
    \centering
    \includegraphics[width=1.\textwidth]{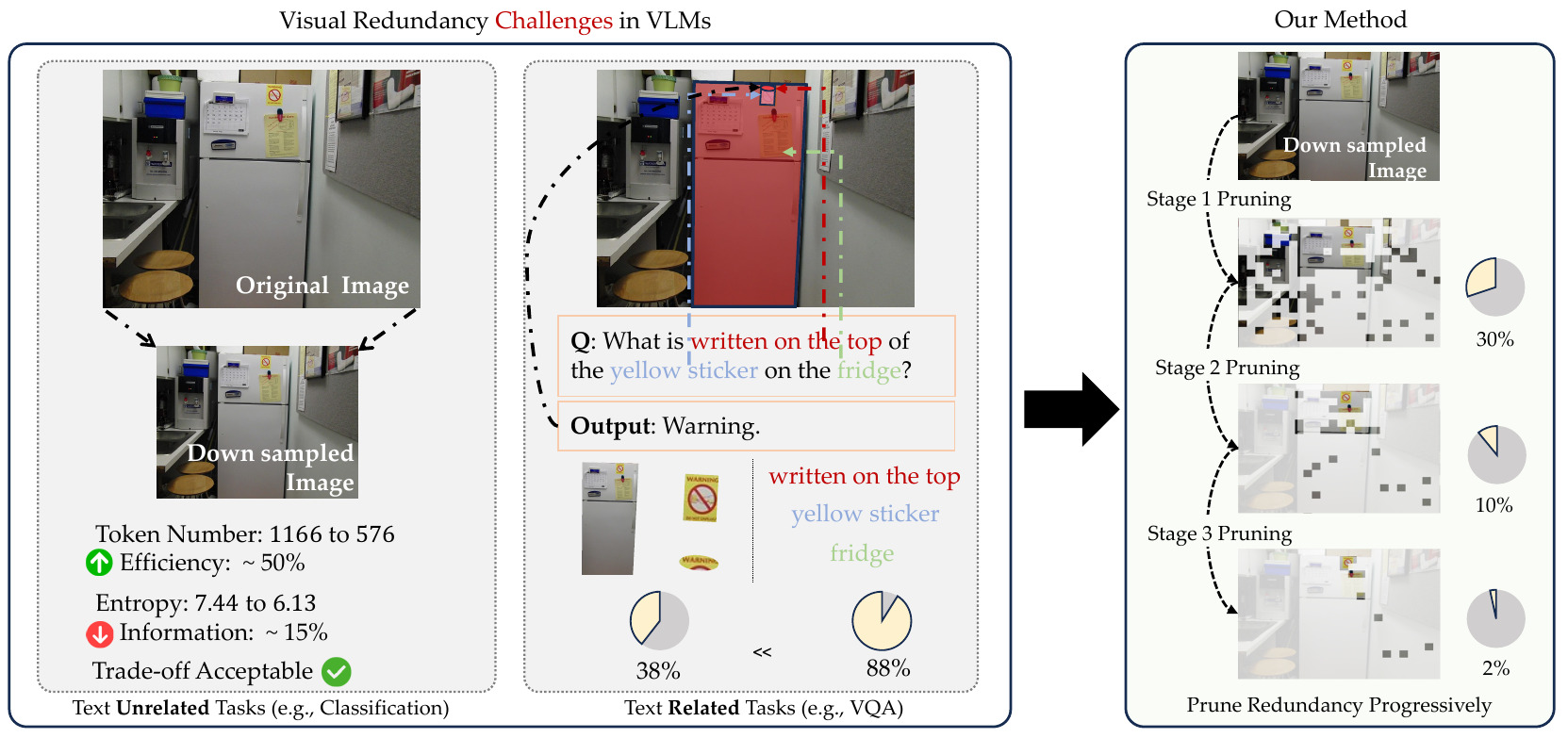}
    \vspace{-6mm}
    \caption{\textbf{Analysis of visual redundancy in different vision tasks.} }
    \label{fig:appendix_moti}
    \vspace{-4mm}
\end{figure*}

In non-textual tasks, such as classification or detection, downsampling is commonly used to reduce visual redundancy and enhance model training efficiency \cite{zhang2024freekd}. Figure \ref{fig:appendix_moti} illustrates this process, showing the reduction of tokens from 1166 to 576 in a downsampled image, resulting in a $50\%$ efficiency boost but a $15\%$ information loss (entropy decreased from 7.44 to 6.13). This trade-off is acceptable for such tasks. Conversely, for text-related tasks like visual question answering (VQA), which involve both text and vision modalities, a distinct approach is required. Highlighting the most information-dense text ($88\%$ of total text) alongside the region pertinent to the query in the image ($38\%$ of total image), we observe that image information is typically sparser than textual data. Hence, our SparseVLM method incrementally prunes visual token redundancy, maintaining crucial information for task accuracy. This strategy enhances model efficiency.

\section{Compatibility with FlashAttention}
\label{a:sparseflashattn}

To ensure compatibility between SparseVLM and FlashAttention \cite{dao2022flashattention} when extracting the matrix $\mA$ or $\mP$, we devise the \textbf{dual-flash attention} operation to directly obtain the average attention scores relative to the text raters. This operation is lightweight and enjoys the efficiency of FlashAttention. Specifically, the first forward pass operates identically to the original FlashAttention, generating the necessary hidden states. In the second forward pass, we introduce a specially designed $\mV$ matrix. In this matrix, for the rows corresponding to the text raters we wish to analyze, we set the values to the reciprocal of the number of text raters. This configuration allows the inner product between the attention map and the $\mV$ matrix to return the mean value of the attention scores for the selected text raters directly in FlashAttention. 
With the mean value, we perform a top-$k$ selection to identify the visual tokens to retain. Tokens that are excluded during this process are converted into masks, which are then applied to the hidden states produced by the first FlashAttention pass to complete the pruning operation. This method enables efficient integration of pruning with FlashAttention while preserving compatibility and computational efficiency. The specific principles and calculation of SparseVLM FlashAttention are as follows:
\begin{enumerate}
\item \textbf{Attention Score Calculation.} For each block $B$, compute the scaled dot-product attention scores as 
\[
\mS_B = \frac{\mQ_B \mK_B^T}{\sqrt{d_k}},
\]
where $\mS_B$ is the attention score matrix computed within the block.

\item \textbf{Block-wise Softmax.} To ensure numerical stability, the Softmax is computed using the log-sum-exp trick as

\begin{enumerate}
    \item Subtract the maximum value for numerical stability:
    \[
    \mS'_B = \mS_B - \max(\mS_B, \text{axis}=1)
    \]
    \item Normalize:
    \[
    \mP_B = \frac{\exp(\mS'_B)}{\sum \exp(\mS'_B, \text{axis}=1)}
    \]
\end{enumerate}

\item \textbf{Special $\mV$ Matrix.} In order to return the mean value of the attention scores for the selected text raters directly with the FlashAttention, we need to design a special $\mV$ matrix.

\[
\mV_{ij} =
\begin{cases}
1/n, & \text{if } i \in \{i_1, i_2, \dots, i_k\}, \\\\
0, & \text{otherwise}.
\end{cases}
\]

Here, $\mV$ is an $n \times d$ matrix, $n$ is the total number of rows in the matrix, $i$ is the row index, $1 \leq i \leq n$, $\bm{s} = \{ i \mid \vr_i \geq m \}, \, i \in \{1, 2, \dots, L_t\}$ define the text raters which we selected in Section 3.2.

\item \textbf{Incremental Accumulation.} Rather than storing $\mP$ explicitly, the result is directly accumulated into the output using:

\[
\mO_B = \mP_B \cdot \mV_B
\]

The final result is obtained by concatenating all blocks:

\[
\mO = \text{Concat}(\mO_1, \mO_2, \ldots, \mO_B)
\]

\item \textbf{Streaming Softmax.} When combining multiple blocks, an incremental softmax computation ensures that normalization is maintained across the entire sequence:

\[
\textrm{Softmax}(\mS) = \exp(\mS) / \sum \exp(\mS)
\]

This avoids global dependencies and enables efficient block-wise computation.

\item \textbf{Top-$k$ Selection for Visual Tokens.} The top-$k$ selection can be expressed as:

\[
\mO_k = \{ x_i \in \mO_v \mid \text{rank}(x_i, \mO_v) \leq k \},
\]

\[
\mO_v = \{ y_j \in \text{mean}(\mO) \mid \text{visual tokens start} \leq j \leq \text{visual tokens end} \}.
\]

where $\mO = \textrm{Concat}(\mO_1, \mO_2, \ldots, \mO_B)$ is the output array of the second FlashAttention, $\mO_v$ is the visual tokens part of $\mO$, $\text{rank}(x_i, \mO_v)$ represents the position of $x_i$ in $\mO_v$ when sorted in descending order.

The corresponding indices of the top-$k$ elements are:

\[
\mI_k = \{ i \mid x_i \in \mO_k \}.
\]

\item \textbf{Summary of SparseVLM with FlashAttention using Top-$k$ Selection.} The complete process of SparseVLM FlashAttention can be summarized as

\[
\begin{aligned}
\mI_k = \{ i \mid x_i \in \{ y_j \in \mO_v \mid 
& \text{rank}(y_j, \text{mean}(\textrm{Concat}\left( \bigcup_{B} 
\text{Softmax}\left(\frac{\mQ_B \mK_B^T}{\sqrt{d_k}} - \max(\mS_B)\right) \cdot \mV_B \right) \\
& [\text{visual tokens start} : \text{visual tokens end}] )) \} \}.
\end{aligned}
\]

Here, each block $B$ is processed independently, and the results are combined using incremental normalization.

\end{enumerate}

\section{Computing Budget Detailed Estimation}
\label{a:flops}
\textbf{Estimation of Visual Token Significance.} 
In this stage, only the \eqref{eq:4} averaging process requires computation. Each visual token undergoes $L_{t}-1$ additions and one division. With $L_v$ visual tokens in total, the number of FLOPs for this stage is $(L_{t}-1 +1) \times L_v = L_t \times L_v$.

\textbf{Relevant Text Selection.} In this process, given that official PyTorch implementation for Softmax and Averaging operations, the FLOPs for \eqref{eq:7} can be approximately simplified to the matrix multiplication between $H_v$ and $H_q$. The result has a shape of $L_v \times L_t$, where each element undergoes $D$ multiplications and additions. Therefore, the FLOP count can be expressed as $L_t \times L_v \times 2D$.

\textbf{Sparsification Level Adaptation.} 
The rank of a matrix is typically computed using singular value decomposition (SVD) \citep{stewart1993early}. With the selected appropriate threshold, the number of above the threshold singular values determines the rank of the matrix. The FLOPs involved in this process can be approximated as $L_t \times L_v \times \text{min}(L_t, L_v)$.

\textbf{Token Aggregation.} 
At this stage, the first part is to perform a nearest neighbor search for each element in the matrix. With the $L_r \times D$ matrix, this task can be simplified to calculate the distances between $L_r$ elements, resulting in a total of $L_r\times(L_r - 1)/2$ distance calculations. Each distance computation requires sequentially executing subtraction, squaring, addition, and square root operations on $D$ elements. Consequently, the number of FLOPs in the nearest neighbor search is $L_r\times(L_r - 1)/2 \times 4D = L_r\times(L_r - 1) \times 2D$.

The second part is density calculation. Since the operations of averaging and applying the exponential function are implemented by the official PyTorch, this part can be simplified by the matrix squaring. Therefore, the FLOPs for this part are $L_r \times L_r \times 2D$.

The third part is distance indicator calculation. The computation can be approximately simplified to compute $\rho_i \times \delta_i$. Therefore, the FLOPs for this part can be approximated as $L_r \times L_r \times 2D$.

The last part is clustering. In this part, we need to select $C$ tokens with the highest scores from a total of $L_r$ tokens to serve as cluster centers, and the FLOPs can be approximated as $L$.

In summary, the total FLOPs for this stage are given by
\label{ta:1}
$$\text{FLOPs} = 
\underbrace{L_r\times(L_r - 1) \times 2D}_{\text{Nearest Neighbors Search}} + \underbrace{L_r \times L_r \times 2D}_{\text{Density Calculation}} + \underbrace{L_r \times L_r \times 2D}_{\text{Distance Indicator Calculation}} + \underbrace{L}_{\text{Select Cluster Center}} 
$$
$$ = L_r\times(3L_r - 1) \times 2D + L.$$

\textbf{Token Reconstruction.} 
Token reconstruction involves performing a weighted sum for each group, excluding the cluster center. Thus, there are $L_r - C$ elements to sum where each one has $1 \times D$ dimensions. Consequently, the number of FLOPs for this operation is $D \times (L_r-C)$.

\section{Dataset}
\label{a:dataset}
We conducted experiments on several widely used visual understanding benchmarks.

\textbf{GQA.}~\citep{hudson2019gqa} The GQA is composed of three parts: scene graphs, questions, and images. The image part contains images, as well as the spatial features of images and the features of all objects in images. The questions in GQA are designed to test the understanding of visual scenes and the ability to reason about different aspects of an image.

\textbf{MMBench.}~\citep{liu2023mmbench} The MMBench benchmark comprehensively evaluates the model’s overall performance across multiple dimensions. It includes three levels of ability dimensions. The first level (L-1) consists of two main abilities, perception and reasoning. The second level (L-2) expands based on the first level, including six sub-abilities. The third level (L-3) further refines the second level, encompassing 20 specific ability dimensions. This hierarchical structure enables a granular and comprehensive evaluation of the model's various capabilities.

\textbf{MME.}~\citep{fu2023mme} The MME benchmark is also a comprehensive benchmark meticulously designed to thoroughly evaluate various aspects of a model's performance. It consists of 14 subtasks that specifically aim to evaluate both the model's perceptual and cognitive abilities. By utilizing manually constructed instruction-answer pairs and concise instruction design, it effectively mitigates issues such as data leakage and unfair evaluation of model performance.

\textbf{POPE.}~\citep{li2023evaluating} The POPE benchmark is primarily used to evaluate the degree of Object Hallucination in models. It reformulates hallucination evaluation by requiring the model to answer a series of specific binary questions regarding the presence of objects in images. Accuracy, Recall, Precision, and F1 Score are effectively employed as reliable evaluation metrics to precisely measure the model's hallucination level under three different sampling strategies.

\textbf{ScienceQA.}~\citep{lu2022learn} The ScienceQA benchmark covers a rich diversity of domains, including natural science, language science, and social science. Within each subject, questions are categorized first by the topic, then by the category, and finally by the skill. This hierarchical categorization results in 26 topics, 127 categories, and 379 skills, providing a comprehensive and diverse range of scientific questions. It provides a comprehensive evaluation of a model's capabilities in multimodal understanding, multi-step reasoning, and interpretability.

\textbf{VQA-v2.}~\citep{goyal2017making} The VQA-v2 benchmark evaluates the model’s visual perception capabilities through open-ended questions. It consists of 265,016 images, covering a wide variety of real-world scenes and objects, providing rich visual contexts for the questions. For each question, there are 10 ground truth answers provided by human annotators, which allows for a comprehensive evaluation of the performance of different models in answering the questions accurately.

\textbf{TextVQA.}~\citep{singh2019towards} The TextVQA benchmark focuses on the comprehensive integration of diverse text information within images. It meticulously evaluates the model’s text understanding and reasoning abilities through a series of visual question-answering tasks with rich textual information. Models need to not only understand the visual content of the images but also be able to read and reason about the text within the images to answer the questions accurately.

\textbf{MMVet.}~\citep{yu2023mm} The MMVet benchmark is designed based on the insight that the intriguing ability to solve complicated tasks is often achieved by a generalist model being able to integrate different core vision-language capabilities. MM-Vet defines 6 core VL capabilities and examines the 16 integrations of interest derived from the capability combination. 

\textbf{TGIF-QA.}~\citep{jang2017tgif} The TGIF-QA benchmark is an extension of the image question answering (ImageQA) task to the video domain, aiming to promote the development of video question answering techniques. It contains 165,000 question answer pairs in total and requires the model to comprehend the details of GIF videos. Specifically, it introduces three new tasks for VideoQA (repetition count, repeating action, and state transition), which require spatio-temporal reasoning from videos, and frame QA tasks that can be answered from one of the frames. 

\textbf{MSVD-QA.}~\citep{xu2017video} The MSVD-QA benchmark is based on the existing Microsoft Research Video Description (MSVD) dataset and contains 1970 video clips and approximately 50.5K QA pairs. The questions and answers are diverse in nature, covering a wide range of topics and aspects related to the video content. Due to its relatively large data size and the diversity of questions, it is widely used for video question answering tasks and video caption tasks. The tasks formed in it are open-ended questions, consisting of five types of questions: what, who, how, when, and where. 

\textbf{MSRVTT-QA.}~\citep{xu2017video} The MSRVTT-QA benchmark consists of 10K video clips and 243k question answer pairs. One of the main challenges addressed by the MSRVTT-QA benchmark is the complexity of understanding and reasoning about video content. Videos contain both visual and temporal information, and models need to be able to effectively process and integrate these aspects to answer the questions accurately. The tasks formed in it also consist of five types of questions, similar to the MSVD-QA benchmark.

\textbf{ActivityNet-QA}~\citep{yu2019activityqa} The ActivityNet-QA benchmark contains 58,000 human-annotated QA pairs on 5,800 videos derived from the ActivityNet dataset. The questions are designed to cover a range of types, including motion, spatial relationship, and temporal relationship, which challenge the model to understand and reason about the video content at different levels and evaluate the performance of VideoQA models in long-term spatio-temporal reasoning.

\section{Implementation Details}
All of our experiments are conducted on a single NVIDIA A100-80G GPU. The implementation is carried out in Python $3.10$, utilizing PyTorch $2.1.2$, CUDA $11.8$, and transformers $4.31.0$. The inference follows the evaluation settings established by LLaVA \citep{liu2024visual}. For LLaVA-1.5-7/13B, Mini-Gemini (MGM), and Qwen-VL, we follow the same inference setting as the original paper as it is publicly available\footnote{\url{github.com/haotian-liu/LLaVA}} \footnote{\url{github.com/dvlab-research/MGM}} \footnote{\url{https://github.com/QwenLM/Qwen-VL}}. For video understanding tasks, we adopt the same inference setup as the original Video-LLaVA code base\footnote{\url{github.com/PKU-YuanGroup/Video-LLaVA}.}, as it is publicly available.

\section{Efficiency Details}
We present a comparative efficiency analysis of SparseVLM, the baseline, and FastV during the inference phase in Table \ref{tab:main}. In this section, we provide additional details on the CUDA time during the inference phase. Following VoCo-LLaMA \cite{ye2024voco}, we primarily consider the following components that contribute to the reported CUDA time: image encoding time (if applicable), KV cache load time (if applicable), and transformers forward time. We exclude other computational times that are not dependent on the model itself and the caching strategy, such as model loading time, from the CUDA time measurement. Specifically, the attention operation is implemented by FlashAttention \cite{dao2022flashattention}.

\section{More Detailed Efficiency Analysis}
\label{a:trade-off}
\begin{figure}[htbp]
    \centering
    \includegraphics[width=0.95\textwidth]{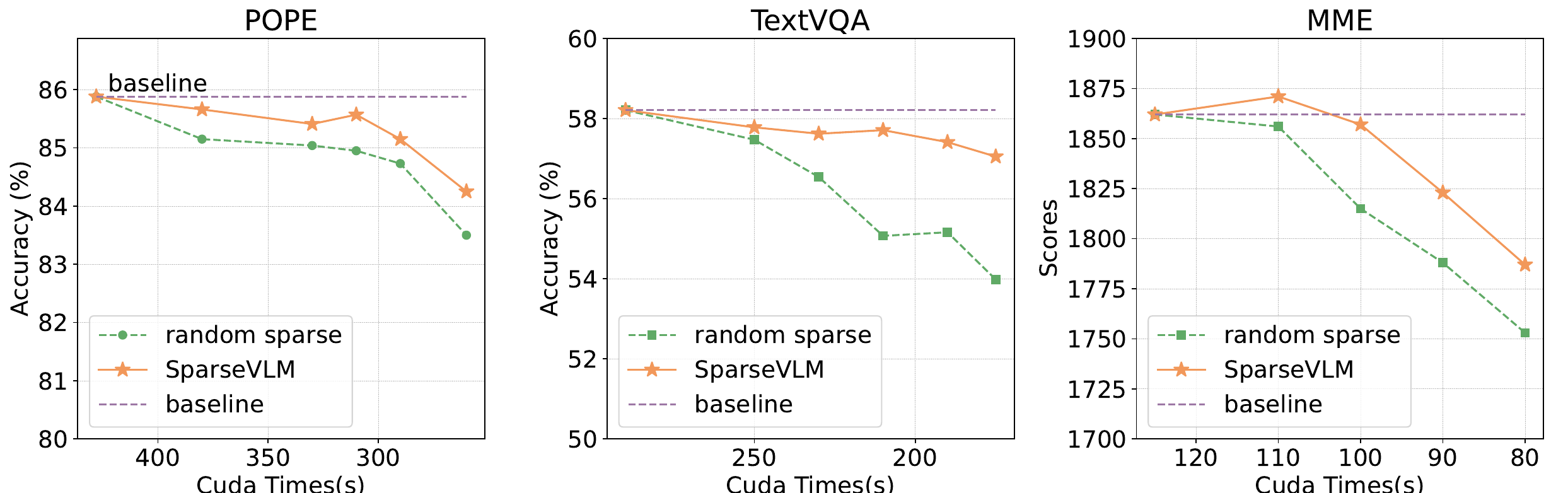}
    \vspace{1em}
    \includegraphics[width=0.95\textwidth]{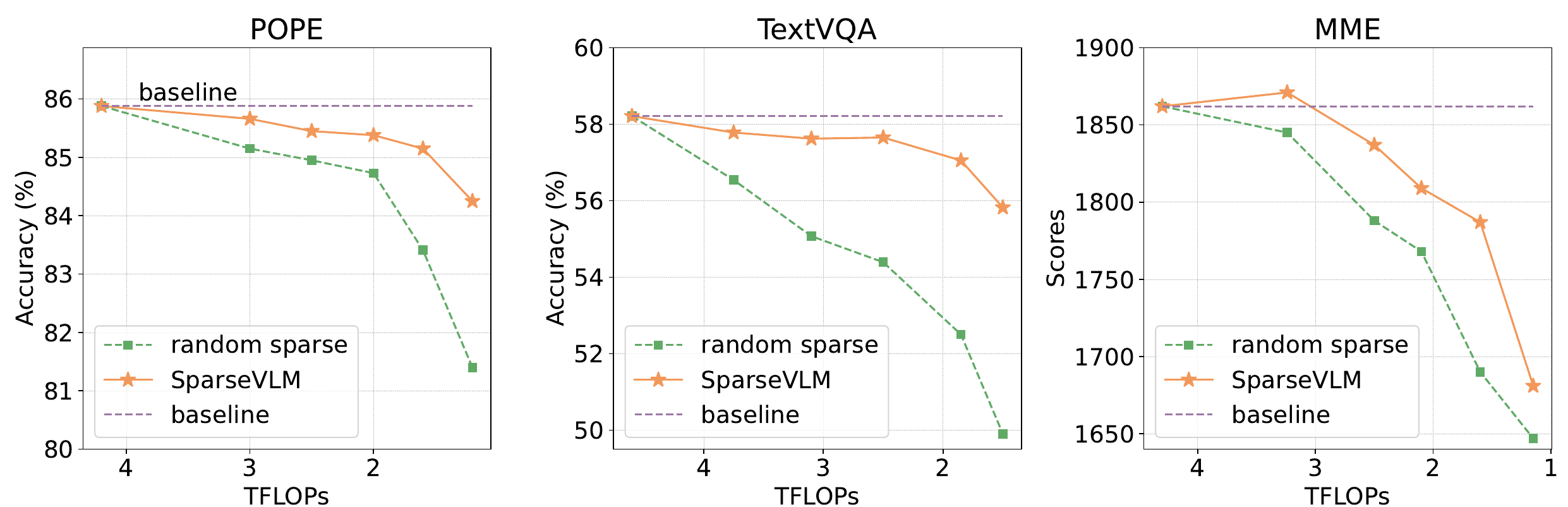}
    \caption{{\textbf{Trade-offs for SparseVLM on LLaVA: (a) Latency vs. Accuracy, and (b) FLOPs vs. Accuracy.} Both studies demonstrate comparisons among random sparse, our SparseVLM, and baseline models.}}
    \label{fig:LLaVA-tradeoff}
\end{figure}

\begin{figure}[htbp]
    \centering
    \includegraphics[width=0.95\textwidth]{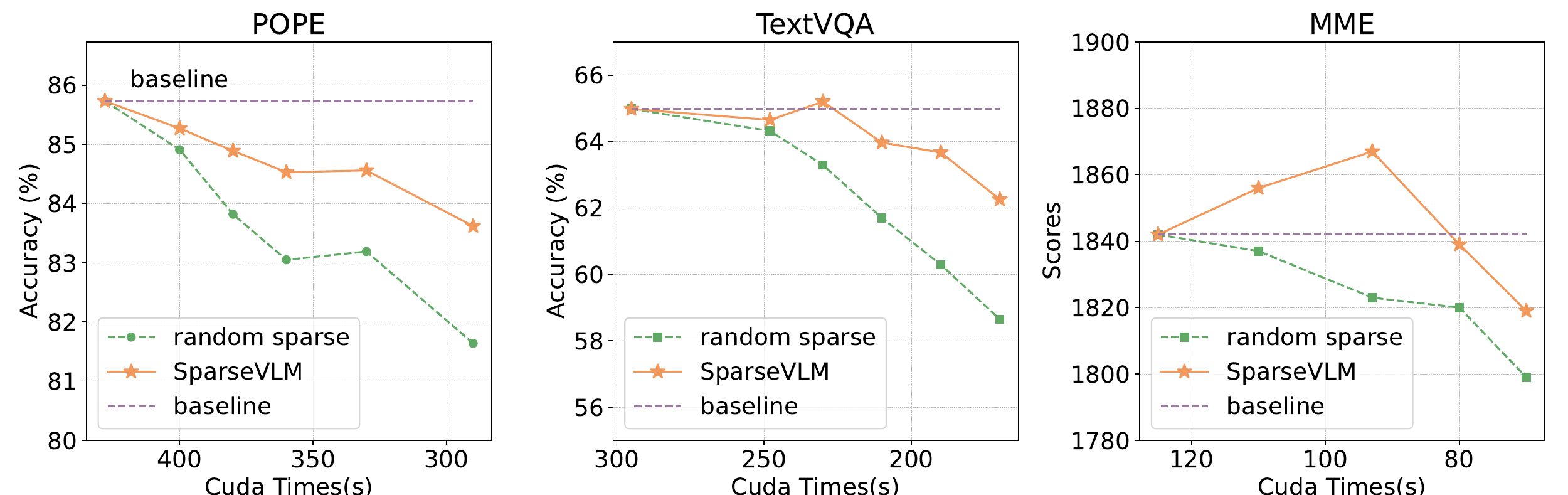}
    \vspace{1em} %
    \includegraphics[width=.95\textwidth]{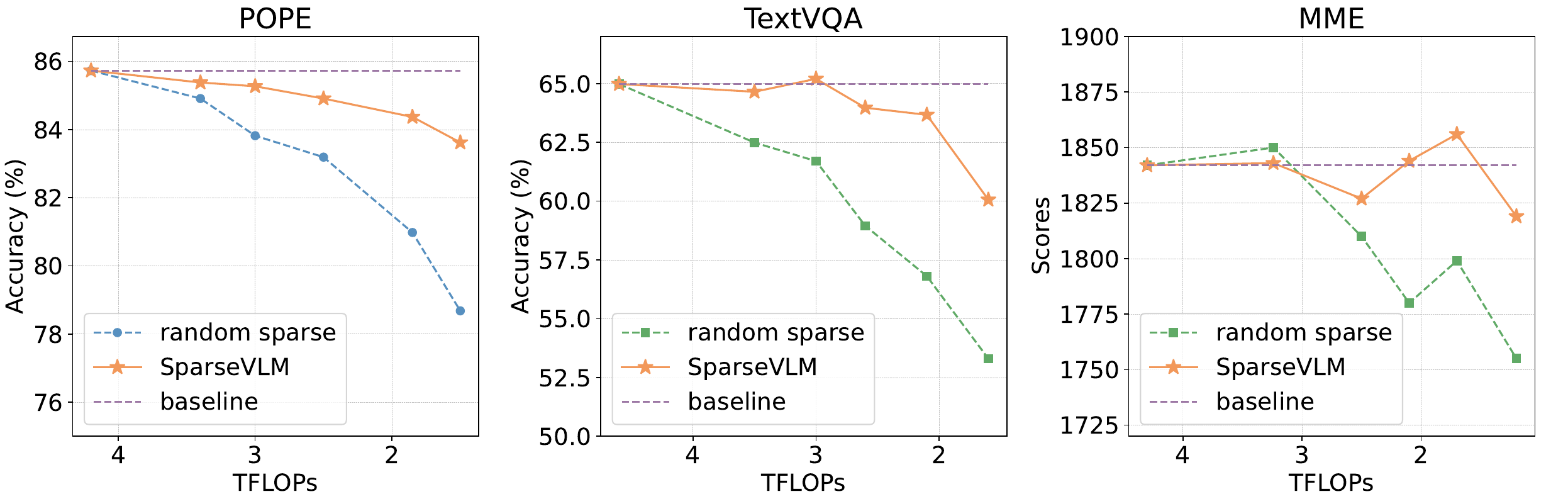}
    \caption{{\textbf{Trade-offs performance for SparseVLM on MGM: (a) Latency vs. Accuracy, and (b) FLOPs vs. Accuracy.} Both studies demonstrate comparisons among random sparse, our SparseVLM, and baseline models.}}
    \label{fig:MGM-tradeoff}
\end{figure}

\begin{figure}[htbp]
    \centering
    \includegraphics[width=0.95\textwidth]{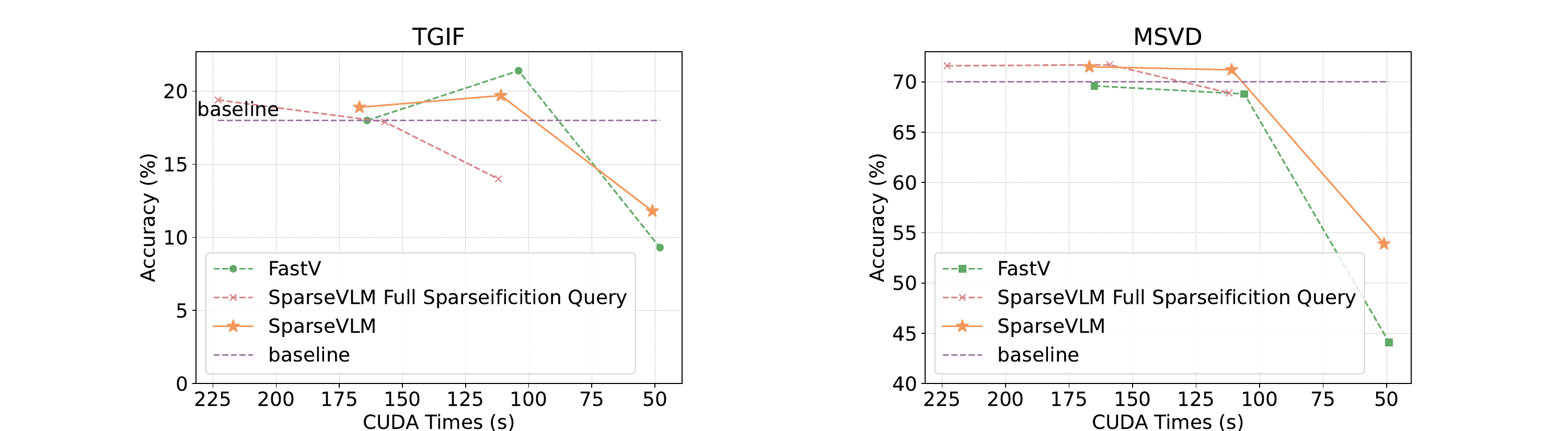}
    \vspace{1em}
    \includegraphics[width=0.95\textwidth]{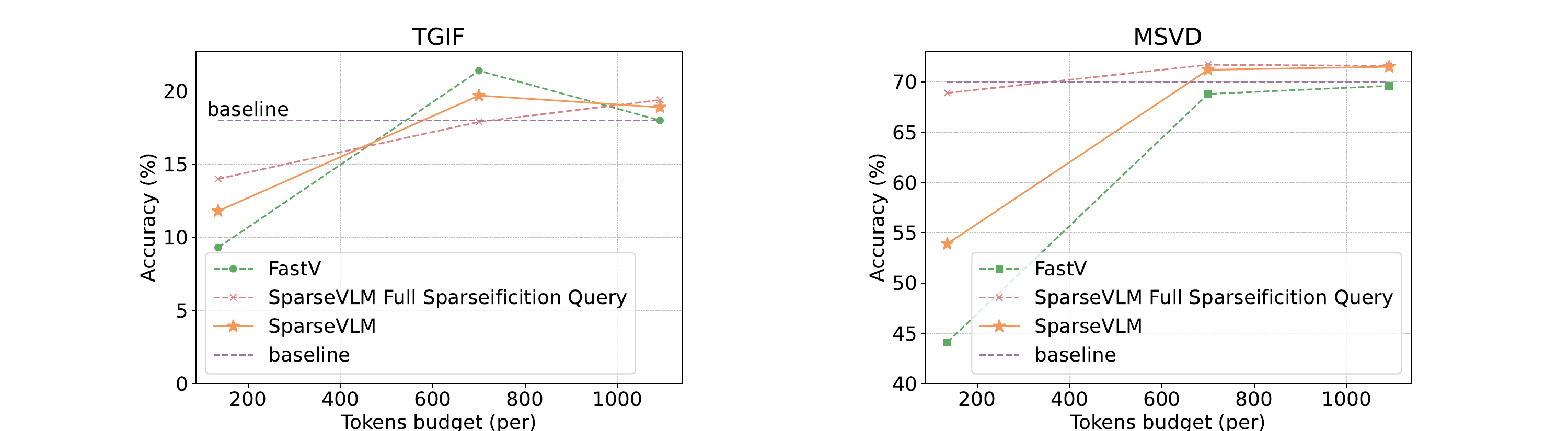}
    \caption{{\textbf{Trade-offs for SparseVLM on Video-LLaVA: (a) Latency vs. Accuracy, and (b) Token budget vs. Accuracy.} Both studies demonstrate comparisons among random sparse, our SparseVLM, and baseline models.}}
    \label{fig:Video-LLaVA-tradeoff}
\end{figure}

To better validate the efficiency of our method, we provide the latency-vs.-accuracy and FLOPs-vs.-Accuracy trade-offs for SparseVLM applied to LLaVA and MGM across three benchmarks: POPE, TextVQA, and MME, which are shown in Figure \ref{fig:LLaVA-tradeoff} and Figure \ref{fig:MGM-tradeoff}. Besides, we also analyze Video-LLaVA matched with SparseVLM in Figure \ref{fig:Video-LLaVA-tradeoff} on TGIF and MSVD.

\section{More Sparsification Visualization}
\label{a:visualization}
\begin{figure*}[t]
    \centering
    \vspace{-2mm}
    \includegraphics[width=.88 \textwidth]{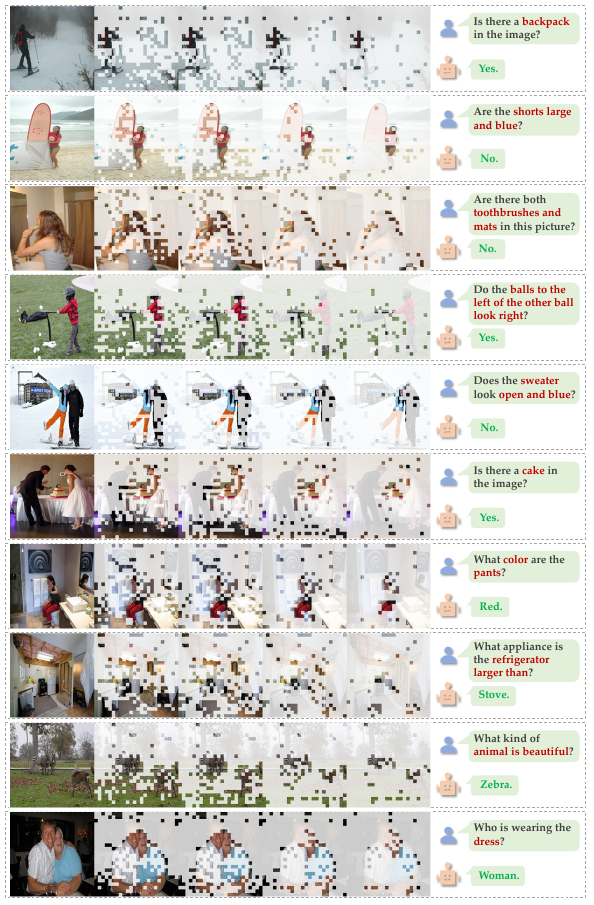}
    \vspace{-4mm}
    \caption{\textbf{More visualization examples of SparseVLM on different prompts.} Best viewed in color.}
    \label{fig:appendix_vis}
    \vspace{-2mm}
\end{figure*}

Figure \ref{fig:appendix_vis} showcases a diverse array of visualization examples that demonstrate the application of SparseVLM across a spectrum of visual question-answering (VQA) prompts. These visualizations offer a deeper insight into how our SparseVLM processes and responds to different types of queries posed in a visual context. 

